\documentclass[journal]{IEEEtran}
\usepackage{amsmath,amsfonts}
\usepackage{array}
\usepackage[caption=false,font=normalsize,labelfont=sf,textfont=sf]{subfig}
\usepackage{textcomp}
\usepackage{stfloats}
\usepackage{url}
\usepackage{verbatim}
\usepackage{graphicx}
\usepackage{cite}
\usepackage{amssymb}
\usepackage{bm}
\usepackage{xcolor}

\usepackage{multirow}
\usepackage{amstext}
\usepackage{algorithm}
\usepackage{algpseudocode}
\usepackage{bbding}
\usepackage[inline]{enumitem}
\usepackage{makecell}
\usepackage{booktabs}
\definecolor{fqr}{RGB}{35,146,144}
\definecolor{fqf}{RGB}{242,70,57}

\def\etal{\emph{et al}.}

\newcommand{\yl}[1]{{{#1}}}
\newcommand{\yyw}[1]{#1}

\usepackage{hyperref}

\begin{document}

\title{OAHuman: Occlusion-Aware 3D Human Reconstruction from Monocular Images}

\author{Yuanwang Yang$^\dagger$, Hongliang Liu$^\dagger$, Muxin Zhang, Nan Ma,~\IEEEmembership{Senior Member,~IEEE,} Jingyu Yang,~\IEEEmembership{Senior Member,~IEEE,} Yu-Kun Lai,~\IEEEmembership{Senior Member,~IEEE,} and Kun Li$^*$, ~\IEEEmembership{Senior Member,~IEEE}
\thanks{$^\dagger$ Equal contribution.}
\thanks{$^*$ Corresponding author: Kun Li (E-mail: lik@tju.edu.cn).}
\thanks{Yuanwang Yang,  Hongliang Liu, Muxin Zhang and Kun Li are with the College of Intelligence and Computing, Tianjin University, Tianjin 300350, China. E-mail: \{yyw,liuhongl,muxinzhang,lik\}@tju.edu.cn}
\thanks{Nan Ma is with the College of Artificial Intelligence, Beijing University of Technology, Beijing 100124, China. E-mail: manan123@bjut.edu.cn}
\thanks{Yu-Kun Lai is with the School of Computer Science and Informatics, Cardiff University, Cardiff CF24 4AG, United Kingdom. E-mail: Yukun.Lai@cs.cardiff.ac.uk}
\thanks{Jingyu Yang is with the School of Electrical and Information Engineering, Tianjin University, Tianjin 300072, China. E-mail: yjy@tju.edu.cn}
}



\maketitle

\begin{abstract}
Monocular 3D human reconstruction in real-world scenarios remains highly challenging due to frequent occlusions from surrounding objects, people, or image truncation. Such occlusions lead to missing geometry and unreliable appearance cues, severely degrading the completeness and realism of reconstructed human models. Although recent neural implicit methods achieve impressive results on clean inputs, they struggle under occlusion due to entangled modeling of shape and texture.
In this paper, we propose \textbf{OAHuman}, an occlusion-aware framework that explicitly decouples geometry reconstruction and texture synthesis for robust 3D human modeling from a single RGB image. 
\yyw{The core innovation lies in the decoupling-perception paradigm, which addresses the fundamental issue of geometry–texture cross-contamination in occluded regions. Our framework ensures that geometry reconstruction is perceptually reinforced even in occluded areas, isolating it from texture interference. In parallel, texture synthesis is learned exclusively from visible regions, preventing texture errors from being transferred to the occluded areas. This decoupling approach enables OAHuman to achieve robust and high-fidelity reconstruction under occlusion, which has been a long-standing challenge in the field.}
Extensive experiments on occlusion-rich benchmarks demonstrate that OAHuman achieves superior performance in terms of structural completeness, surface detail, and texture realism, significantly improving monocular 3D human reconstruction under occlusion conditions. The source code will be available for research purposes.
\end{abstract}


\section{Introduction}
\label{sec:intro}

Capturing a complete and unobstructed image of the entire human body is often impractical in real-world scenarios. Whether in sports events, group activities, or everyday street scenes, human bodies are frequently occluded by other people, environmental objects, or suffer from truncation due to a limited camera field of view. Such occlusions lead to missing structural and appearance information, significantly limiting the performance of monocular 3D human reconstruction methods in practical applications.  

Recent advances in neural implicit representations~\cite{saito2019pifu,saito2020pifuhd,xiu2023econ,qiu2025lhm} have enabled high-quality human reconstruction from a single image. However, these methods typically assume clean, fully visible inputs. When occlusions occur, they often fail to recover complete geometry or produce consistent textures, as shape and appearance are jointly modeled and mutually entangled. 

\yyw{Occlusions introduce two fundamental challenges. (1) They disrupt body structure, making geometric inference in invisible regions highly uncertain. (2) They remove reliable appearance information, leading to unrealistic and inconsistent texture synthesis.
To address these issues, prior studies have explored two main directions: parametric template-based approaches~\cite{cheng2019occlusion,zhang2020object} and implicit neural modeling~\cite{xiang2023wild2avatar}. Parametric methods leverage strong body priors to guide occluded reconstruction but struggle with loose clothing, complex poses, and large deviations from the template. Implicit methods offer greater flexibility but often couple geometry and texture in shared latent spaces, causing error propagation between the two domains. For instance, Wang~\etal~\cite{wang2023complete} use volumetric feature generation to hallucinate plausible body structures. \yl{While} effective in coarse completion, their results lack high-frequency geometric details and suffer from blurred or mismatched textures in occluded areas.  
Despite these efforts, a unified solution that explicitly decouples geometric reasoning from appearance synthesis under occlusion remains underexplored.}

To overcome these limitations, we propose \textbf{OAHuman}, an occlusion-aware monocular 3D human reconstruction framework that explicitly disentangles geometry and texture modeling through a decoupling-perception paradigm. Instead of jointly optimizing shape and appearance—which often leads to mutual interference—OAHuman reconstructs geometry and synthesizes texture in two perceptually guided stages.  
\yyw{This design transforms occlusion handling from passive completion into an active perception problem, where geometric inference and texture synthesis are guided by distinct yet complementary cues of visibility, semantics, and normal consistency.}

In the geometry modeling stage, we introduce a visibility-guided coarse completion module that transfers geometric cues from visible to occluded regions, producing an initial complete body. To ensure semantic coherence, a feature-level supervisor regularization aligns intermediate features with those of a pre-trained reconstruction network, enhancing deep structural consistency. Furthermore, a dual-view normal refinement strategy predicts detailed surface normals from both front and back views, effectively recovering fine-grained geometry even under severe occlusions.  
\yyw{Together, these modules form a coarse-to-fine geometric reasoning pipeline that progressively refines structure from global topology to local surface detail.}

In the texture modeling stage, we design a geometry-faithful diffusion renderer conditioned on the refined normals and image-adaptive style embeddings. This renderer learns texture generation only from visible regions and uses geometric priors to propagate appearance consistently to occluded areas. The result is high-fidelity, view-consistent texture synthesis that respects geometric boundaries and suppresses appearance artifacts. 

\begin{figure*}
    \includegraphics[width=1.0\textwidth]{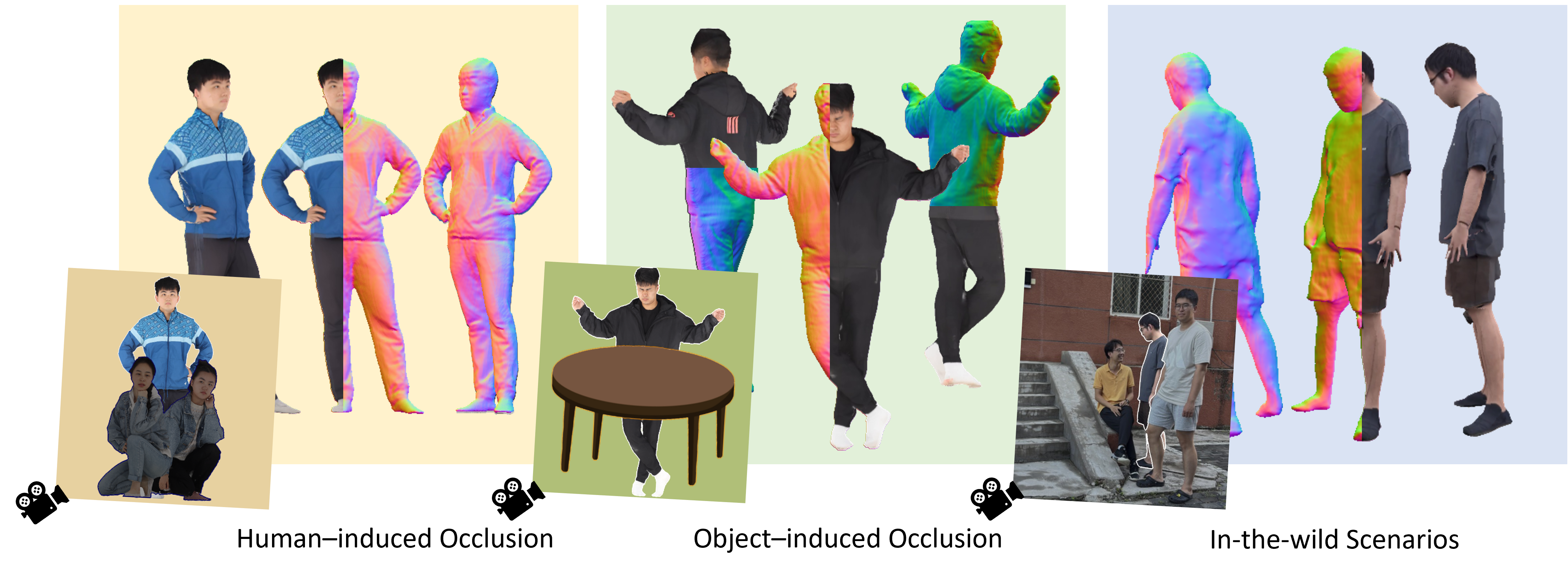}
    \caption{OAHuman enables robust 3D human reconstruction under diverse occlusions. Our method successfully reconstructs complete geometry and realistic textures from monocular images across challenging occlusion types, including human-induced (left), object-induced (middle), and in-the-wild scenarios (right). Input highlights are added for visualization to indicate the target subject.}
    \label{fig:teaser}
\end{figure*}

\yyw{With this decoupling–perception framework, OAHuman breaks the long-standing loop of “geometry corruption–texture confusion” that limits robust reconstruction under occlusion. The framework not only restores complete human geometry but also generates spatially consistent textures aligned with the recovered surface, pushing single-view 3D human reconstruction toward real-world readiness.}
Extensive experiments on occlusion-rich benchmarks demonstrate that OAHuman achieves superior robustness, structural completeness, and texture realism across diverse occlusion conditions. Figure~\ref{fig:teaser} illustrates its ability to reconstruct complete and visually coherent 3D humans even under severe occlusion.

\vspace{0.5em}
\noindent{Our main contributions are summarized as follows:}
\begin{itemize}
  \item We propose OAHuman, a novel occlusion-aware monocular 3D human reconstruction framework that decouples geometry completion and texture synthesis, improving modeling capacity in occluded regions.
  \item We design a coarse-to-fine geometry reconstruction pipeline combining a visibility-guided coarse completion module with feature-level supervisor regularization and a dual-view normal refinement strategy, enhancing both structural integrity and surface detail recovery.
  \item We introduce a geometry-faithful diffusion renderer that generates high-fidelity and view-consistent textures under normal supervision, leveraging both geometric priors and large-scale generative models.
\end{itemize}

\section{Related Work}

\subsection{Monocular Clothed Human Reconstruction} Monocular 3D reconstruction of clothed humans aims to recover the 3D structure (e.g., mesh) of clothed human bodies from a single image. In this field, implicit functions have become the mainstream technical approach due to their strong adaptability to complex topological structures and diverse clothing styles. Since the pioneering work PIFU~\cite{saito2019pifu} introduced implicit functions into 3D human reconstruction, a series of derivative methods, including Geo-PIFu~\cite{he2020geo}, ICON~\cite{xiu2022icon}, IntegratedPIFu~\cite{chan2022integratedpifu}, ECON~\cite{xiu2023econ}, and the work in~\cite{zhang2023global}, have achieved reconstruction by learning implicit functions representing the enclosed surface of clothed humans, with the core goal of improving the accuracy and stability of geometric modeling. To further enhance reconstruction quality, researchers have introduced auxiliary information from multiple dimensions: PIFUHD~\cite{saito2020pifuhd} optimizes local details by incorporating normal information; PaMIR~\cite{zheng2021pamir} integrates parametric models (such as SMPL) as conditional constraints into implicit representations, improving the accuracy and stability of geometric modeling; FOF~\cite{feng2022fof} proposes a geometric representation method based on Fourier occupancy fields, significantly accelerating reconstruction speed; Learning visibility field~\cite{zheng2023learning} leverages depth cues to enhance geometric consistency in dynamic scenes. Although these methods have developed in diverse directions, they mainly target complete monocular human images, and their performance degrades significantly when dealing with occluded humans.

\subsection{Human Appearance Reconstruction} Methods such as Animate Anyone~\cite{hu2024animate}, Champ~\cite{zhu2024champ}, and OmniHuman~\cite{lin2025omnihuman} can generate high-quality 2D human appearances, but they rely on the completeness of the input image and may suffer from body pose distortions due to the lack of 3D structure. Appearance reconstruction techniques extended from geometric reconstruction have shown breakthroughs in multiple directions in recent years:  R${^2}$Human~\cite{yang2024r} combines the advantages of implicit texture fields and explicit neural rendering, and proposes an intermediate representation Z-map to achieve real-time monocular human reconstruction, but it still suffers from texture blurring or geometric distortion in scenes with severe occlusions or loose clothing. In explicit reconstruction, SiTH~\cite{ho2024sith} uses an image-conditioned diffusion model to generate back views that are perceptually consistent with the front view, complementing the appearance information of unobserved regions. However, it relies on a \verb+"+front-back view sandwich\verb+"+ reconstruction approach, leading to insufficient recovery of geometric and texture details in side views, making it difficult to capture unique clothing folds or contours on the sides. SIFU~\cite{zhang2024sifu} decouples side-view features through a cross-attention mechanism, with normals of the SMPL-X model as queries, improving the accuracy and robustness of 2D-to-3D mapping, and uses text-to-image diffusion priors to generate consistent textures for invisible regions, but its handling of extreme viewpoints or severe occlusions still has room for improvement. PSHuman~\cite{li2024pshuman} proposes a cross-scale multi-view diffusion model to jointly model global full-body and local facial features to avoid distortions, and then recovers textured realistic human meshes from the generated multi-view images through explicit sculpting, but its handling of extreme scenarios (such as severe occlusions) is limited. In implicit reconstruction, Human-LRM~\cite{weng2024template} uses an enhanced geometric decoder to generate tri-plane NeRF (Neural Radiance Field) and then hallucinates multi-view details through a diffusion model, significantly improving the realism and cross-view consistency of occluded regions, but it may have detail deviations in extreme poses/complex clothing due to reliance on diffusion priors. IDOL~\cite{zhuang2025idol} adopts a feed-forward Transformer architecture, combining 3D Gaussian splatting with the SMPL-X parametric model to achieve real-time reconstruction of 3D humans from a single image, but its reconstruction effect on half-body images is poor. LHM~\cite{qiu2025lhm} uses 3D Gaussian splatting as the representation, fuses 3D geometric and 2D image features through a multimodal Transformer, and combines head feature pyramid encoding to enhance facial details, but it has defects in occluded scenarios and poor geometric performance.

\begin{figure*}[t]
  \centering
   \includegraphics[width=0.98\linewidth]{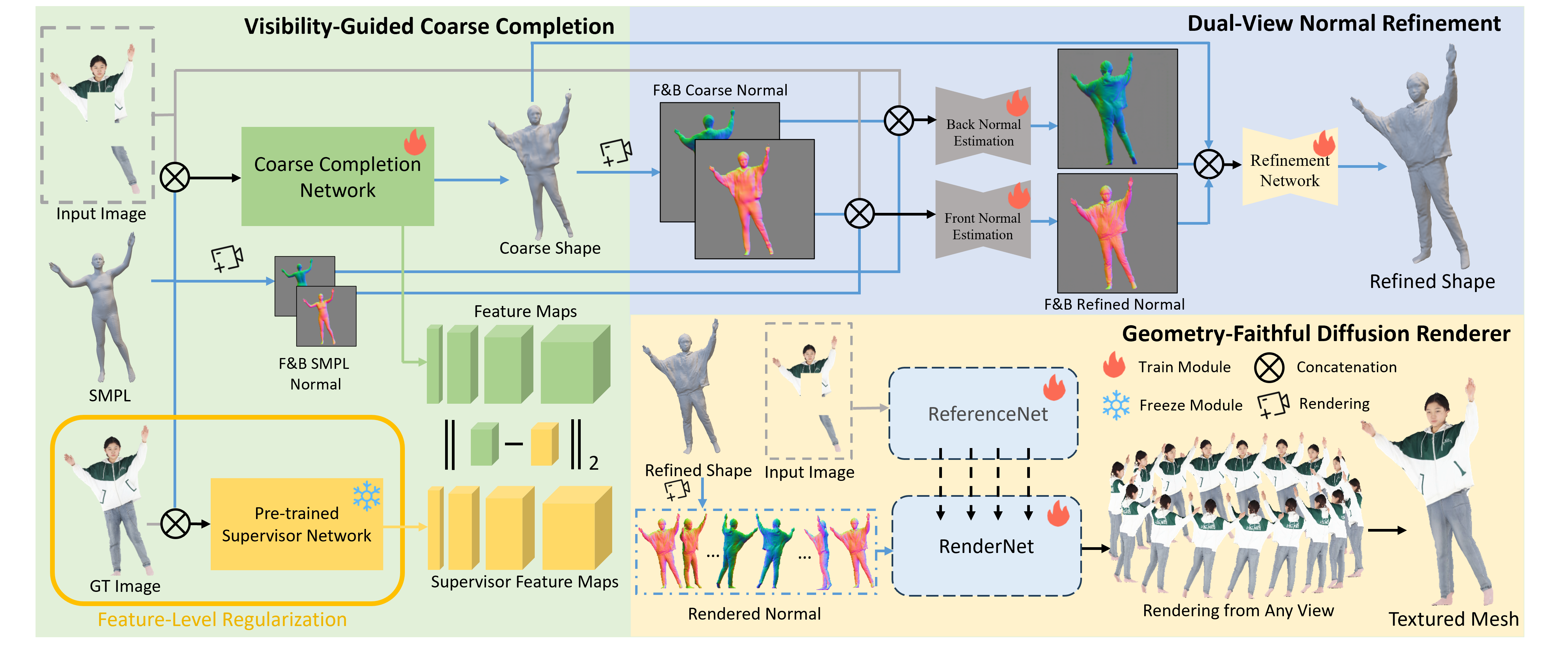}
   \caption{\textbf{Overview of OAHuman.} Our method consists of a two-stage pipeline for occlusion-aware 3D human reconstruction. The first stage performs geometry restoration using a visibility-guided coarse completion (VGCC) module with feature-level supervisor regularization (FLSR) and dual-view normal refinement (DVNR). The second stage synthesizes semantically consistent textures via a geometry-faithful diffusion renderer, enabling high-fidelity reconstruction under occlusion.}
   \label{fig:pipeline}
\end{figure*}

\noindent\subsection{Occlusion Handling in Human Reconstruction}

Occlusion remains a fundamental challenge in human reconstruction and rendering, and has been studied extensively as reconstruction tasks become increasingly complex. Early work primarily focused on handling occlusions in pose and shape estimation. For example, previous studies~\cite{cheng2019occlusion, zhang2020object, xiang2023rendering} investigated the influence of object occlusions on human pose and shape estimation, reducing ambiguity through semantic segmentation and prior constraints. Liu \etal~\cite{liu2022explicit} further modeled occlusion relationships between people for multi-person pose estimation in crowded scenes.

As reconstruction tasks extend from pose estimation to full 3D geometry recovery, researchers have begun to address occlusion in more complex interaction scenarios. Huang \etal~\cite{huang2024closely, huang2025reconstructing} studied reconstruction under close human interactions by incorporating proxemics and physical constraints to reason about inter-person occlusions. Khirodkar \etal~\cite{khirodkar2022occluded} proposed a dedicated mesh recovery scheme to handle severe inter-human occlusions. However, these approaches remain largely constrained to parametric body models such as SMPL and thus have limited capability in recovering detailed surface geometry.

Recent deep learning approaches have further improved occlusion robustness in monocular human reconstruction. CompleteHuman~\cite{wang2023complete} enhances local geometric detail using multi-view normal fusion and progressive texture completion, though it struggles with loose clothing and complex hairstyles. Wild2Avatar~\cite{xiang2023wild2avatar} addresses occluded humans in in-the-wild monocular videos by decoupling humans from occluders through occlusion-aware scene parameterization with SMPL priors, but its reconstruction quality can degrade under heavy occlusion and segmentation errors. Robust-PIFu~\cite{chanrobust} removes external occluders via a disentangling diffusion model and resolves internal occlusions using multi-layer normal prediction, although generative components may introduce hallucinated structures. SCHOR~\cite{dutta2025joint} jointly models scene and human-object interactions using diffusion-based completion, but it relies on predefined object templates and may lack fine details for small structures such as fingers or hair. CHROME~\cite{dutta2025chromelittle} explores occlusion recovery via image-based completion prior to reconstruction; however, its results are limited by low-resolution inputs and often exhibit noticeable artifacts in facial regions.

In contrast to the aforementioned approaches, which struggle with severe occlusions and fine details, in this paper we propose OAHuman, an occlusion-aware framework that decouples geometry and texture synthesis, enhancing reconstruction accuracy and realism in occluded regions.

\section{Method}

Our method aims to reconstruct complete and high-fidelity 3D clothed humans from monocular RGB images containing partial occlusion. To address the fundamental challenges of missing structure and unreliable appearance in occluded regions, we propose a decoupled framework named \textbf{OAHuman}, which consists of two stages: geometry reconstruction and texture synthesis. As illustrated in Figure~\ref{fig:pipeline}, we first restore complete geometry using a visibility-guided coarse completion (VGCC) module with feature-level supervisor regularization (FLSR) and a dual-view normal refinement (DVNR), then synthesize semantically consistent textures using a geometry-faithful diffusion renderer. This decoupled design ensures robust modeling under occlusion by explicitly leveraging structural priors and disentangling geometry from appearance.

\subsection{Visibility-Guided Coarse Completion}

We adopt the Fourier Occupancy Field (FOF)~\cite{feng2022fof} as our base representation, which compresses 3D volumetric occupancy into a 2D Fourier field $\mathbf{C}$ aligned with the input image. Given an occluded image $\mathbf{I}$, we estimate the coefficients via:
\begin{equation}
\mathbf{C} = \varepsilon(\mathbf{I}), \quad O(\mathbf{x}_p, z) = \mathbf{b}(z)^\top \mathbf{C}(\mathbf{x}_p),
\end{equation}
where $\varepsilon$ is a CNN encoder that predicts the 2D Fourier field $\mathbf{C}$, $\mathbf{b}(z)$ is the Fourier basis at depth $z$, $x_p$ is the pixel-aligned 2D location and $O$ denotes the occupancy at the 3D point. This formulation allows efficient regression of a continuous 3D occupancy field $O \in \mathbf{R}^3$, from which we extract the iso-surface at $O=0.5$.

However, because of missing pixels in the occluded regions, the network struggles to recover the complete geometry solely from the image. To compensate for missing visual cues in the occluded regions, we incorporate structural priors from a fitted SMPL mesh $\mathbf{S}$. The mesh is projected into the Fourier space and concatenated with the image:
\begin{equation}
\mathbf{C} = \varepsilon\left([\mathbf{I}, f(\mathbf{S})]\right),
\end{equation}
where $f(\cdot)$ encodes $\mathbf{S}$ into FOF space. This fusion provides coarse anatomical guidance.

We supervise $\mathbf{C}$ using a mean square loss:
\begin{equation}
\mathcal{L}_{\text{mse}} = \sum_i \| \mathbf{C}_i - \mathbf{C}_i^{gt} \|_2^2.
\end{equation}

\begin{figure}[htb]
    \centering
    \includegraphics[width=0.5\textwidth]{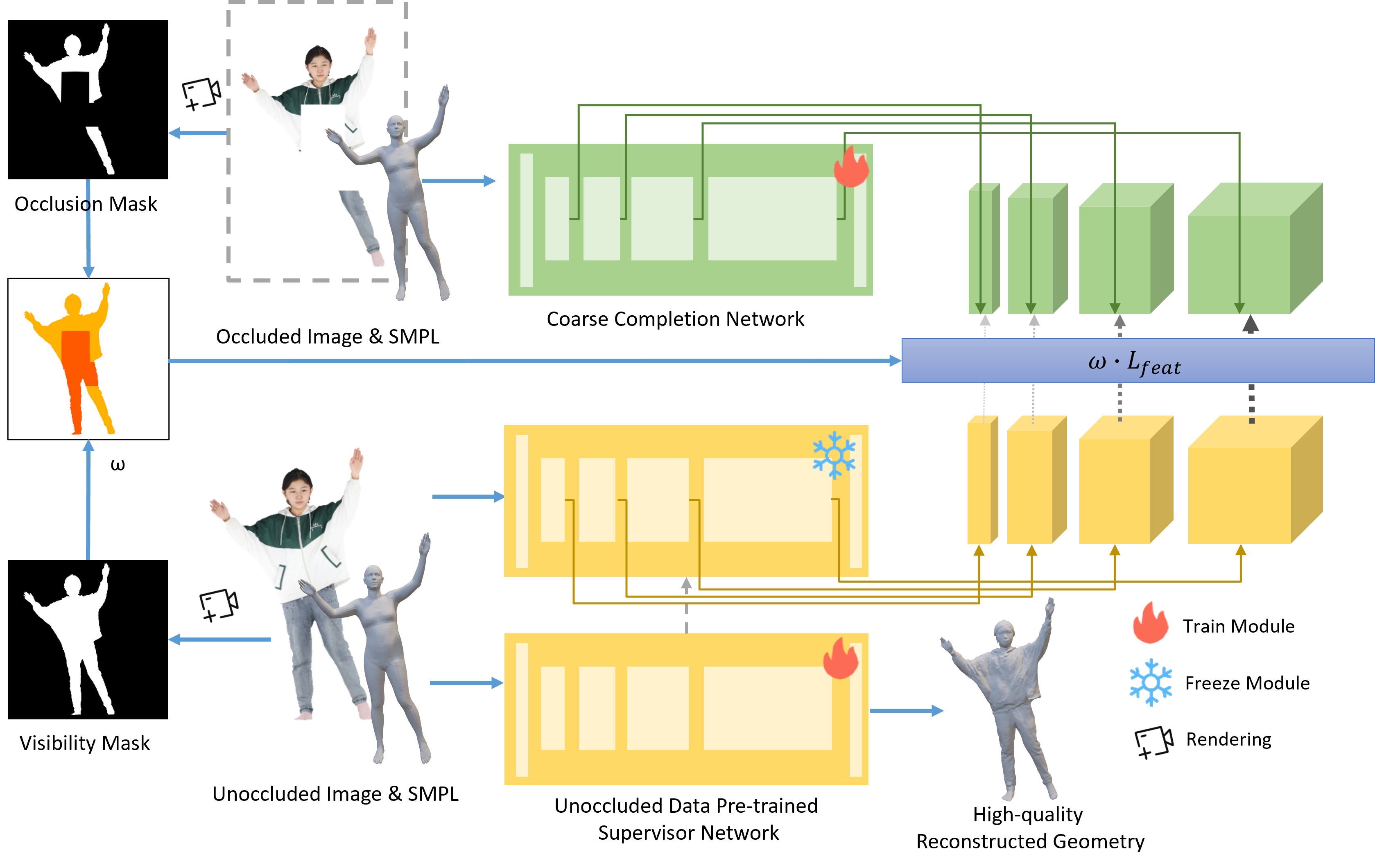}
    \caption{Feature-level supervisor regularization. We use a supervisor network trained on unoccluded data to provide hierarchical feature guidance for geometry completion. We achieve emphasis on global structure and suppression of low-level noise through depth-aware feature weighting, and enhance supervision on occluded body regions through a visibility-aware masking strategy.}
    \label{fig:regular}
\end{figure}

While the parametric prior aids structural plausibility, it still struggles to recover fine-scale geometry in the absence of reliable image features. To further enhance geometry completeness and mitigate hallucinations, we design a \textbf{feature-level supervisor regularization}. As shown in Figure~\ref{fig:regular}, we use a supervisor network $\mathcal{T}$ trained on clean, unoccluded data to provide reference features. Let $F_k$ and $F_k^T$ be features from the current and supervisor networks at semantic level $k$, then:
\begin{equation}
\mathcal{L}_{\text{feat}} = \sum_{k=1}^K w_k \| F_k - F_k^T \|_2^2,
\end{equation}
where $w_k$ is a monotonically increasing weight ($w_1 < \dots < w_K$), emphasizing deep features that encode global shape and suppressing shallow textures that are often unreliable in occluded regions.

To spatially focus the supervision, we introduce a visibility-aware weighting scheme using a visibility map $V \in \{0, 1\}^{H \times W}$ and an occlusion mask $M \in \{0, 1\}^{H \times W}$. The per-pixel supervision weight $\omega(x)$ is defined as:
\begin{equation}
\omega(x) =
\begin{cases}
\lambda_{\text{occ}}, & \text{if } M(x)=1 \\
\lambda_{\text{vis}}, & \text{if } V(x)=1 \text{ and } M(x)=0. \\
0, & \text{otherwise}
\end{cases}
\end{equation}
This design ensures that supervision is concentrated on the human body region, with a higher emphasis on occluded areas and no penalty on background or ambiguous regions.

The total loss for geometry completion is defined as:
\begin{equation}
\mathcal{L}_{\text{geo}} = \mathcal{L}_{\text{mse}} + \lambda \sum_{x} \omega(x) \mathcal{L}_{\text{feat}}(x).
\end{equation}

This visibility-guided, layer-weighted feature supervision enables our model to generate geometrically plausible and semantically coherent shapes even under heavy occlusion.

\subsection{Dual-View Normal Refinement}

Although the coarse model produces a plausible overall shape, it often lacks high-frequency details like cloth wrinkles and surface discontinuities. To enhance surface realism, we propose a dual-view normal refinement strategy using front and back normal maps.

Instead of directly predicting normals from the image (which is unreliable under occlusion), we render coarse normals $\mathbf{N}_f$ and $\mathbf{N}_b$ from the coarse mesh and refine them with learned convolutional modules $G_f$ and $G_b$:
\begin{equation}
\mathbf{N}_f' = G_f\left([\mathbf{I}, \mathbf{N}_f]\right), \quad \mathbf{N}_b' = G_b\left([\mathbf{I}, \mathbf{N}_b]\right).
\end{equation}
Here, the image provides appearance clues while the coarse normals serve as structural initialization. This enables the network to recover consistent high-resolution normals even in occluded areas.

We then combine the refined normals and intermediate geometry feature $\mathbf{C}$ as input to a refinement network $e_r$, which outputs a more accurate reconstruction:
\begin{equation}
\mathbf{C}_r = \varepsilon_r\left([\mathbf{N}_f', \mathbf{C}, \mathbf{N}_b']\right).
\end{equation}
We optimize with a combination of $\mathcal{L}_1$ loss and perceptual loss (LPIPS) to capture both low-frequency shape and high-frequency detail:
\begin{equation}
\mathcal{L}_{\text{normal}} = \mathcal{L}_1(\mathbf{N}, \mathbf{N}^{gt}) + \mathcal{L}_{\text{LPIPS}}(\mathbf{N}, \mathbf{N}^{gt}).
\end{equation}
This module enhances surface fidelity and realism, especially for loose clothing and fine wrinkles.

\begin{figure*}[t]
  \centering
   \includegraphics[width=0.97\linewidth]{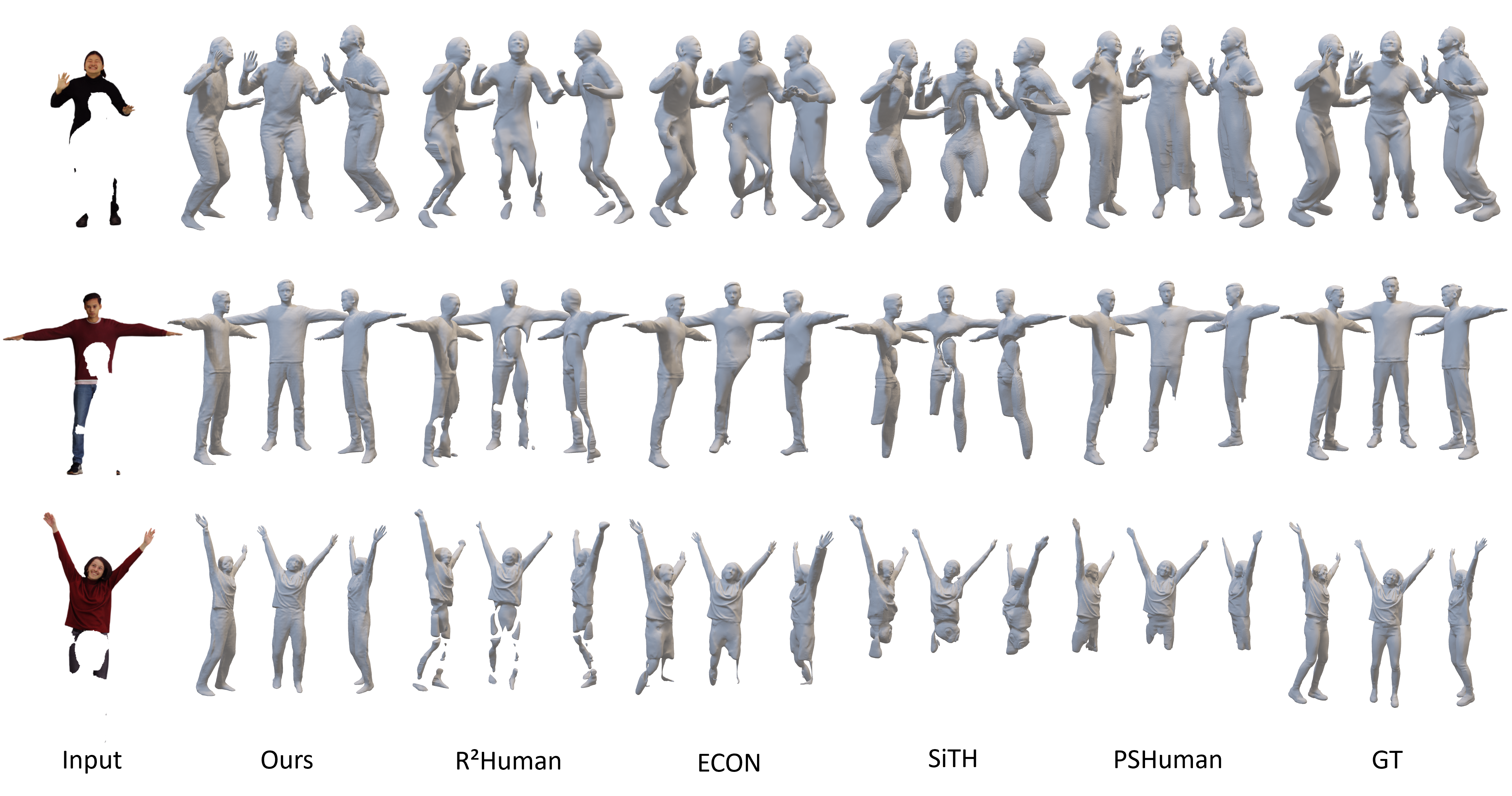}
   \caption{Qualitative comparison of geometry reconstruction results.}
   \label{fig:comp1}
\end{figure*}

\begin{figure*}[t]
  \centering
   \includegraphics[width=0.95\linewidth]{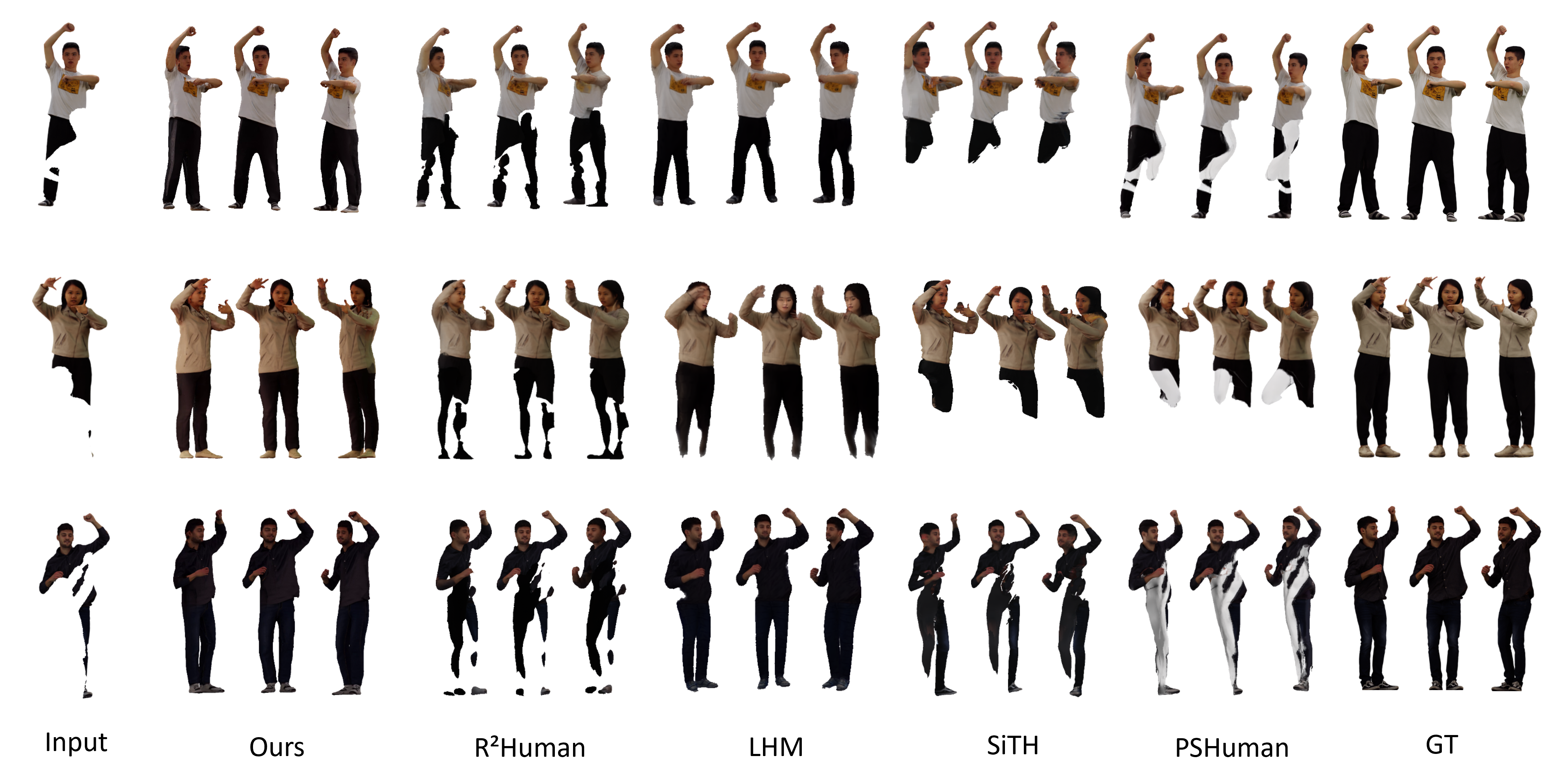}
   \caption{Qualitative comparison of novel view synthesis results.}
   \label{fig:comp2}
\end{figure*}

\subsection{Geometry-Faithful Diffusion Renderer}

To recover spatially consistent and stylistically natural textures from reconstructed geometry, we propose a geometry-aware diffusion rendering framework. Inspired by the dual-branch design in AnimateAnyone~\cite{hu2024animate}, our method introduces a collaborative structure consisting of a main rendering network RenderNet and a style reference network ReferenceNet, while explicitly leveraging 3D structural information to enhance spatial consistency and visual realism.

In contrast to AnimateAnyone, which relies on 2D keypoint-driven synthesis, we incorporate 3D geometry explicitly into the generation pipeline. In RenderNet, we condition the diffusion model on a \emph{ground-truth normal map} \( N \in \mathbb{R}^{H \times W \times 3} \), rendered from the 3D mesh, to guide structure-consistent texture synthesis. The generation process is formulated as:

\begin{equation}
\hat{\textbf{I}} = \text{RenderNet}(N, \eta),
\end{equation}

\noindent where \( \eta \sim \mathcal{N}(0, I) \) denotes Gaussian noise in the diffusion process, and \( \hat{I} \in \mathbb{R}^{H \times W \times 3} \) is the generated texture image. Conditioning on normal maps provides strong structural constraints, leading to better edge continuity and multi-view consistency, particularly in high-curvature regions such as the head, limbs, and garment folds.

To further improve stylistic consistency, we introduce ReferenceNet, which extracts a style vector \( F_s \) from the input occluded image $\textbf{I} \in \mathbb{R}^{H \times W \times 3} $:

\begin{equation}
F_s = \text{ReferenceNet}(\textbf{I}).
\end{equation}

The style vector \( F_s \) is injected into RenderNet's decoder $D$ via a feature fusion mechanism:

\begin{equation}
D_{\text{RenderNet}} \leftarrow \text{Fuse}(F_s, \cdot).
\end{equation}

\begin{table*}[h]
    \centering
    \caption{Quantitative comparison of geometry reconstruction and novel view synthesis.}
    \resizebox{.97\textwidth}{!}{%
    \begin{tabular}{c|cccccc|cccccc}
        \toprule[1.5pt]
        \multirow{3}{*}{Method} 
        &\multicolumn{6}{c}{CustomHumans~\cite{cus}} 
        &\multicolumn{6}{c}{2k2k~\cite{han2023high}}\\
        \cmidrule{2-13}
        &\multicolumn{3}{c}{Geometry} &\multicolumn{3}{c}{Novel View Synthesis} 
        &\multicolumn{3}{c}{Geometry} &\multicolumn{3}{c}{Novel View Synthesis}\\ 
        \cmidrule(lr){2-4}\cmidrule(lr){5-7}
        \cmidrule(lr){8-10}\cmidrule(lr){11-13}
                    &CD$\downarrow$   &P2S$\downarrow$ &Normal$\downarrow$ &SSIM$\uparrow$   &PSNR$\uparrow$  &LPIPS$\downarrow$
                    &CD$\downarrow$   &P2S$\downarrow$ &Normal$\downarrow$ &SSIM$\uparrow$   &PSNR$\uparrow$  &LPIPS$\downarrow$\\ 
        \hline
        ECON         &1.9223 &1.2334 &1.5150                 
                     &-      &-      &- 
                     &2.8937 &2.6024 &3.1966                 
                     &-      &-      &- \\
        SiTH         &4.2868 &1.8328 &2.0597                 
                     &0.8698 &23.08  &0.1261 
                     &4.1096 &3.2313 &3.0603
                     &0.8706 &15.07  &0.1448 \\
        R${^2}$Human &1.6711 &1.5694 &1.6085                 
                     &0.8331 &21.84  &0.1981 
                     &2.9063 &2.5428 &2.4865                 
                     &0.8325 &15.53  &0.1168 \\
        PSHuman      &3.0263 &2.5027 &2.7213
                     &0.8257 &18.85  &0.4894
                     &3.2485 &2.6949 &2.7534
                     &0.8838 &15.99  &0.1142\\
        LHM          &-      &-      &-                 
                     &0.8336 &20.95  &0.1718
                     &-      &-      &-            
                     &0.8416 &14.94  &0.1277\\
        Ours         
        &\textbf{0.9435} &\textbf{0.9010} &\textbf{1.1624}
        &\textbf{0.8759} &\textbf{23.51 } &\textbf{0.0833}
        &\textbf{1.4501} &\textbf{1.3437} &\textbf{1.4603}
        &\textbf{0.9093} &\textbf{22.48 } &\textbf{0.0593}\\
        \bottomrule[1.5pt]
    \end{tabular}%
    }
    \label{tab:1}
\end{table*}

We also incorporate a geometry consistency module inspired by AnimateAnyone's temporal alignment mechanism. Instead of 2D keypoint-based regularization, we apply 3D normal alignment across views to ensure consistent geometry under occlusions and pose variations. This constraint effectively suppresses texture misalignments and visual artifacts, especially in occluded or ambiguous regions.

During training, we optimize the renderer using a combination of pixel-level and perceptual-level objectives:

\begin{equation}
\mathcal{L}_{\text{render}} = \lambda_1 \| \hat{\textbf{I}} - \textbf{I}_{\text{gt}} \|_1 + \lambda_2 \mathcal{L}_{\text{LPIPS}}(\hat{\textbf{I}}, \textbf{I}_{\text{gt}}),
\end{equation}

\noindent where $\textbf{I}_{\text{gt}}$  denotes the ground-truth image, and $\mathcal{L}_{\text{LPIPS}}$  measures perceptual similarity in deep feature space.

In summary, our geometry-faithful diffusion renderer combines accurate 3D normals with style priors from occluded inputs, and incorporates a 3D-guided consistency constraint. This enables high-fidelity, view-consistent human texture synthesis from monocular images, even in the presence of significant occlusions and pose variation.

\section{Experiments}

\subsection{Comparison}

We evaluate our method against several recent state-of-the-art monocular 3D human reconstruction approaches, including ECON~\cite{xiu2023econ}, SiTH~\cite{ho2024sith}, R$^2$Human~\cite{r2human}, LHM~\cite{qiu2025lhm}, and PSHuman~\cite{li2024pshuman} (comparison with~\cite{wang2023complete} is precluded by unavailable implementation). Following the training protocol of ECON and SiTH, we train our model on the publicly available Thuman2.0 dataset~\cite{tao2021function4d}. To further assess generalization capability and ensure consistent evaluation, \yyw{we conduct comparative experiments on the CustomHumans dataset~\cite{cus} and the 2k2k dataset~\cite{han2023high}}, as some previous methods do not release official training/testing splits.

For evaluation, we adopt both 3D and 2D metrics to comprehensively assess reconstruction quality. Specifically, we report Chamfer Distance (CD), point-to-surface distance (P2S), and normal map error to measure geometric accuracy, while LPIPS, SSIM, and PSNR are used to evaluate perceptual similarity and image fidelity of the rendered results. We benchmark 3D reconstruction errors against ground-truth meshes, and use multi-view rendering from frontal, dorsal, and side perspectives to evaluate texture quality.

As shown in Table~\ref{tab:1}, our method achieves the best performance across all evaluation metrics. Notably, ECON lacks full-body texture generation and LHM uses implicit rendering without explicit geometry, limiting their performance. Our method reduces CD and P2S errors by a large margin compared to previous approaches, and also achieves lower normal map error, indicating improved surface detail recovery. On the image side, we obtain higher SSIM and PSNR scores, along with significantly lower LPIPS, reflecting the high-fidelity and perceptual realism of the synthesized textures. 

Qualitative results further support these findings. As shown in Figure~\ref{fig:comp1}, our method reconstructs more complete and structurally accurate geometry under occlusion, while Figure~\ref{fig:comp2} demonstrates visually coherent texture synthesis from novel views. In contrast, baseline methods often suffer from artifacts in the occluded regions, with incomplete geometry or texture holes. Our method benefits from visibility-guided geometry completion and normal-aware texture synthesis, enabling accurate and photorealistic 3D human reconstruction even from severely occluded inputs.

\begin{table}[htb]
    \centering
    \caption{Quantitative results of ablation study.}
    \resizebox{.45\textwidth}{!}{%
    \begin{tabular}{c|ccc}
    \toprule[1.5pt]
    Method &CD$\downarrow$ &P2S$\downarrow$ &Normal$\downarrow$\\ 
    \hline
    VGCC w/o FLSR   &1.4283 &1.4281  &1.4551\\
    VGCC      &1.2547 &1.2275  &1.3717\\
    2D Normal Completion         &0.9835 &0.9408  &1.2124\\
    \hline
    Ours(full) &\textbf{0.9440}&\textbf{0.8831}&\textbf{1.1082}\\
    \bottomrule[1.5pt]
    \end{tabular}%
    }
    \label{tab:2}
\end{table}

\subsection{Ablation Study}

\subsubsection{Ablation on Geometry Completion}
To evaluate the contribution of each component in our pipeline, we conduct ablation studies with several model variants:
(1) \textbf{VGCC w/o FLSR}: a baseline that performs coarse geometry completion using only 2D-3D fused features, without feature-level supervision regularization;
(2) \textbf{VGCC}: This variant introduces feature-level supervisor regularization (FLSR), which guides the completion process using semantic priors from a supervisor network;  
(3) \textbf{2D Normal Completion}: instead of volumetric geometry recovery, this variant directly predicts front and back normal maps from the occluded image, and reconstructs the surface via SMPL-based fusion;  
(4) \textbf{Ours}: the complete model, combining feature-level supervision regularization and Dual-View Normal Refinement (DVNR) to recover both global shape and high-frequency details.

\begin{figure}[htb]
    \centering
    \includegraphics[width=0.4\textwidth]{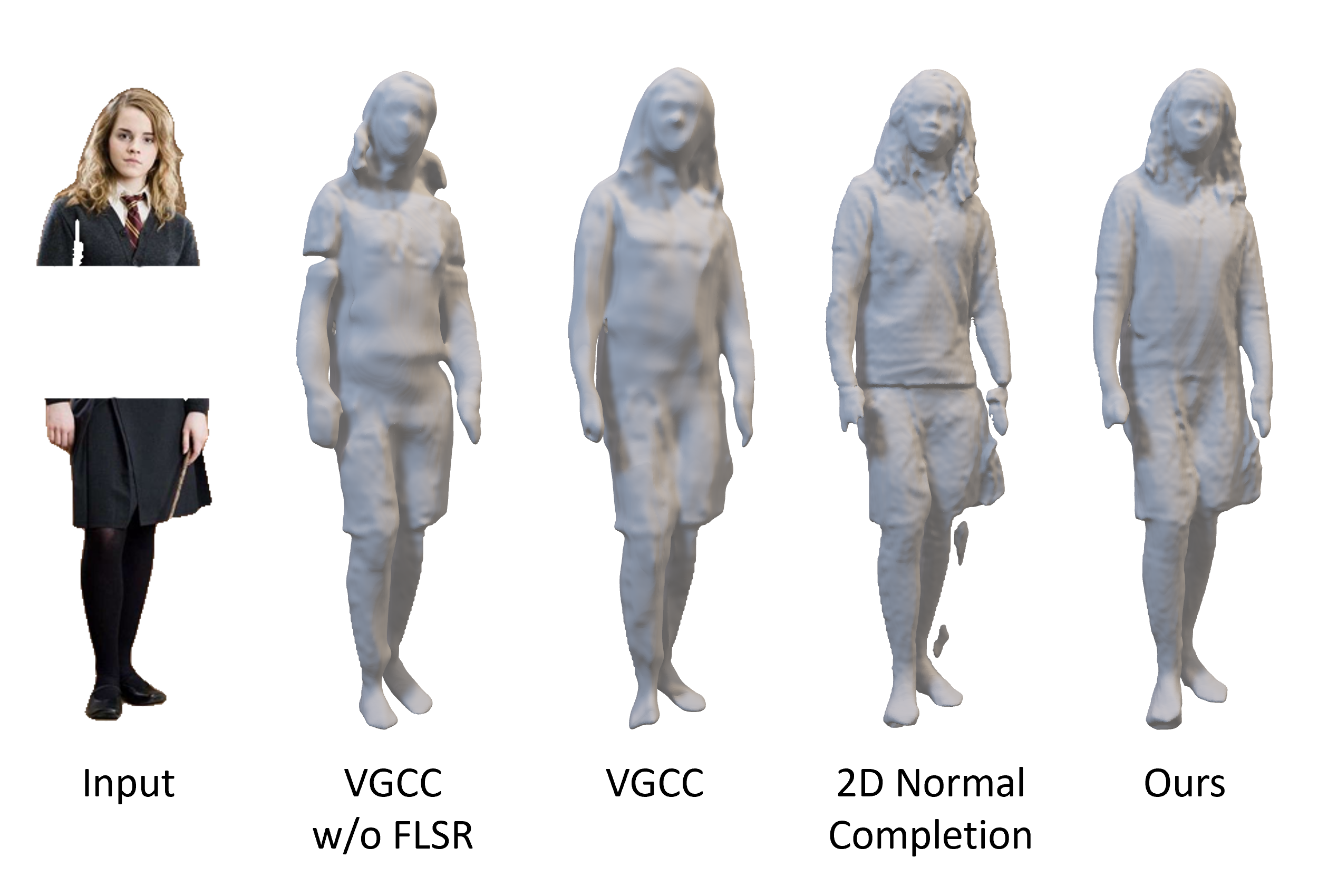}
    \caption{Qualitative results of ablation study.}
    \label{fig:ablation}
\end{figure}

As shown in Table~\ref{tab:2}, the basic VGCC w/o FLSR model recovers the overall human shape but often produces discontinuities at joint boundaries and heavily occluded regions, resulting in fragmented surfaces (see Figure~\ref{fig:ablation}). Introducing FLSR brings clear improvements by leveraging semantic guidance to enforce geometric consistency and completeness.

However, fine-scale details such as surface curvature and clothing wrinkles remain underrepresented. By integrating DVNR, our full model further enhances local fidelity, accurately capturing high-frequency geometry. In contrast, the 2D normal completion variant suffers from artifacts like floating patches and inconsistent surfaces, due to the absence of volumetric constraints. These results underscore the importance of 3D-aware geometry reasoning in occluded settings.

\begin{table}[ht]
    \centering
    \caption{Ablation on the renderer with different normal sources.}
    \resizebox{.45\textwidth}{!}{%
    \begin{tabular}{c|ccc}
    \toprule
    Method & SSIM$\uparrow$ & PSNR$\uparrow$ & LPIPS$\downarrow$\\ 
    \hline
    w/o Mesh normal   &0.8672 &22.45  &0.0891\\
    w/ Mesh normal &\textbf{0.8759} &\textbf{23.51} &\textbf{0.0833}\\
    \bottomrule
    \end{tabular}}
    \label{tab:3}
\end{table}

\begin{figure}[htb]
  \centering
   \includegraphics[width=1\linewidth]{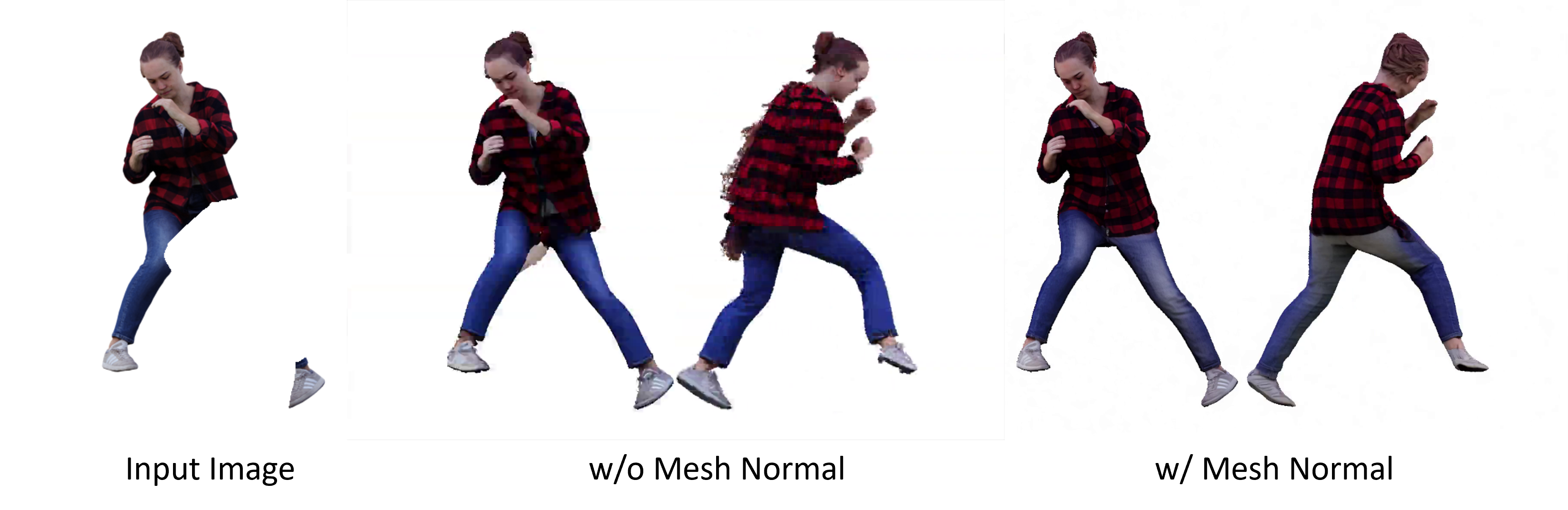}
   \caption{Qualitative comparison of renderer ablations. Using refined mesh normals leads to more faithful texture generation.}
   \label{fig:alb}
\end{figure}

\subsubsection{Ablation on the Renderer}

We further evaluate the impact of different normal sources in our geometry-faithful diffusion renderer. Specifically, we compare conditioning the renderer on SMPL normals (w/o Mesh normal) versus using refined mesh normals derived from our completed geometry (w/ Mesh normal).

As shown in Table~\ref{tab:3}, replacing SMPL normals with refined mesh normals consistently improves all texture metrics, achieving higher SSIM and PSNR and lower LPIPS. Figure~\ref{fig:alb} qualitatively illustrates the rendering improvements. Conditioning on refined mesh normals provides more accurate local surface orientation, enabling the diffusion renderer to better preserve geometric structures and produce view-consistent textures.
\begin{figure*}[t]
  \centering
   \includegraphics[width=.8\linewidth]{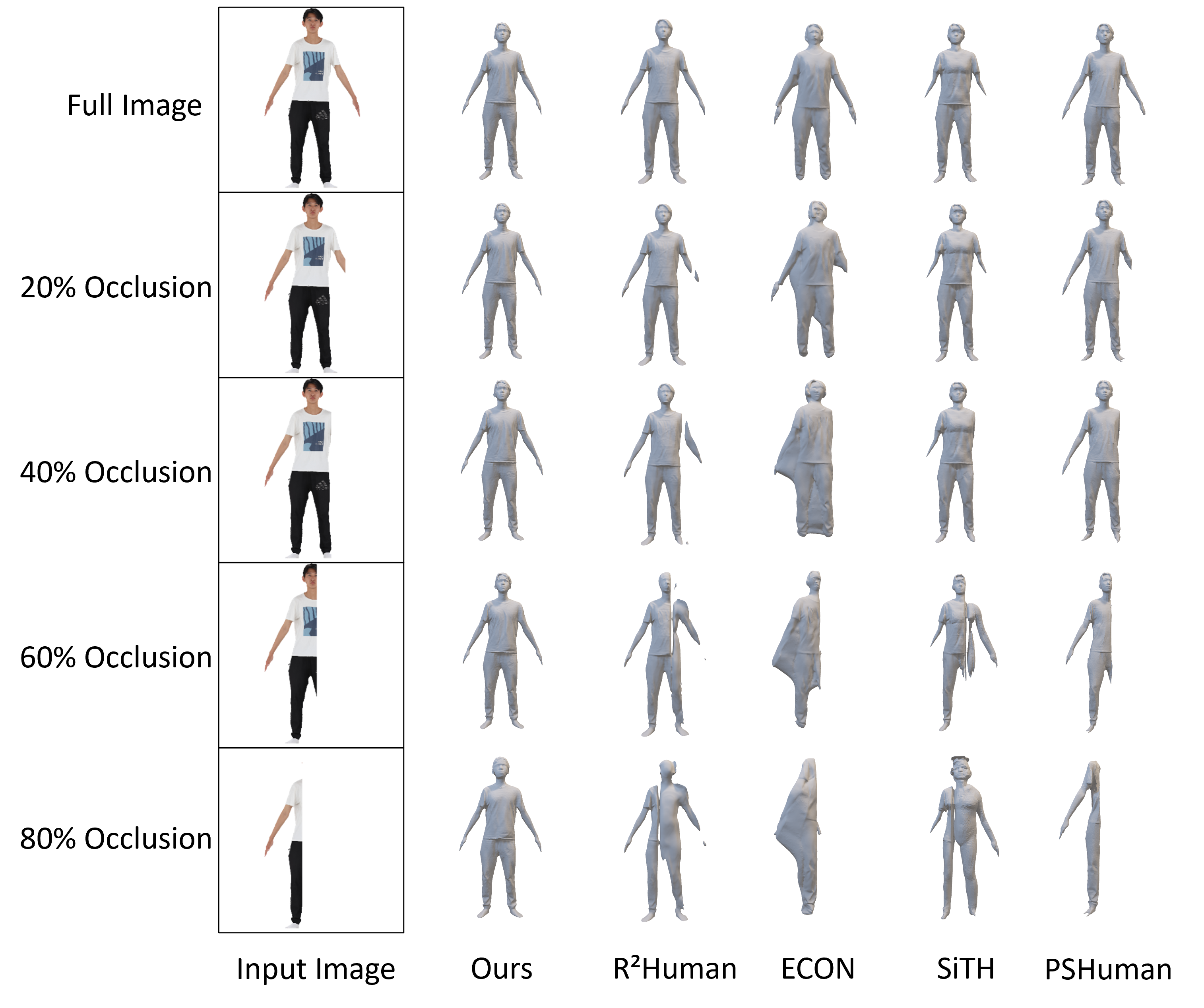}
   \caption{Qualitative reconstruction results under accumulative occlusion.}
   \label{fig:occ}
\end{figure*}
\begin{figure}[t]
  \centering
   \includegraphics[width=1\linewidth]{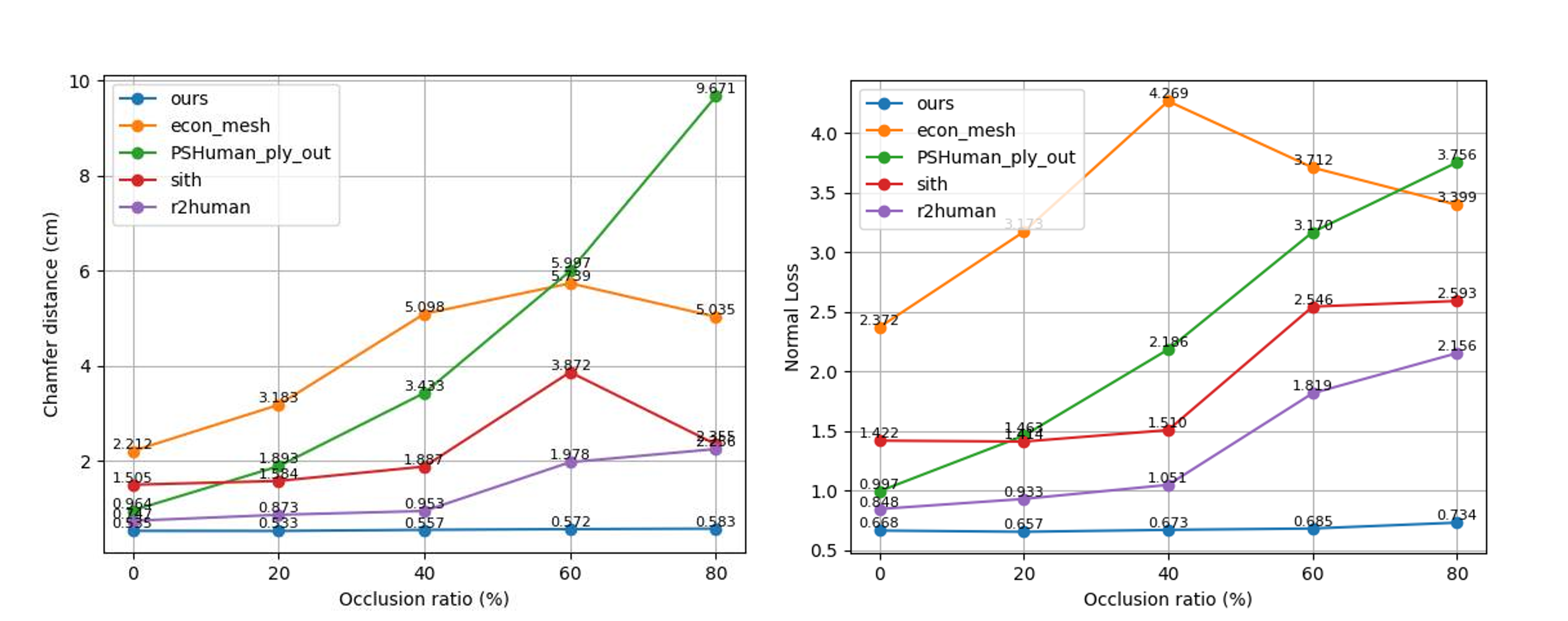}
   \caption{Occlusion-to-accuracy quantitative curves.}
   \label{fig:occ_num}
\end{figure}

\subsection{Robustness Analysis}
\subsubsection{Impact of Occlusion Level on Reconstruction}
To analyze robustness under increasing occlusion, we evaluate reconstruction performance across different occlusion ratios.

Figure~\ref{fig:occ_num} presents quantitative curves relating occlusion level to reconstruction accuracy. 
Chamfer Distance is used to measure geometric reconstruction quality, while normal loss evaluates surface fidelity.

As the occlusion ratio increases from 20\% to 80\%, the performance of existing methods such as ECON~\cite{xiu2023econ} and PSHuman~\cite{li2024pshuman} degrades significantly. 
R$^2$Human~\cite{r2human} and SiTH~\cite{ho2024sith} frequently produce broken limbs or discontinuities near occlusion boundaries.

In contrast, our approach maintains stable reconstruction quality across all occlusion levels. 
The visibility-guided geometry completion allows the model to infer missing body structures even under severe occlusion, leading to more coherent global geometry.

Qualitative comparisons in Fig.~\ref{fig:occ} further demonstrate that our method produces consistent human shapes while baseline methods suffer from structural artifacts when large regions are occluded.
\begin{figure}[ht]
\includegraphics[width=.95\linewidth]{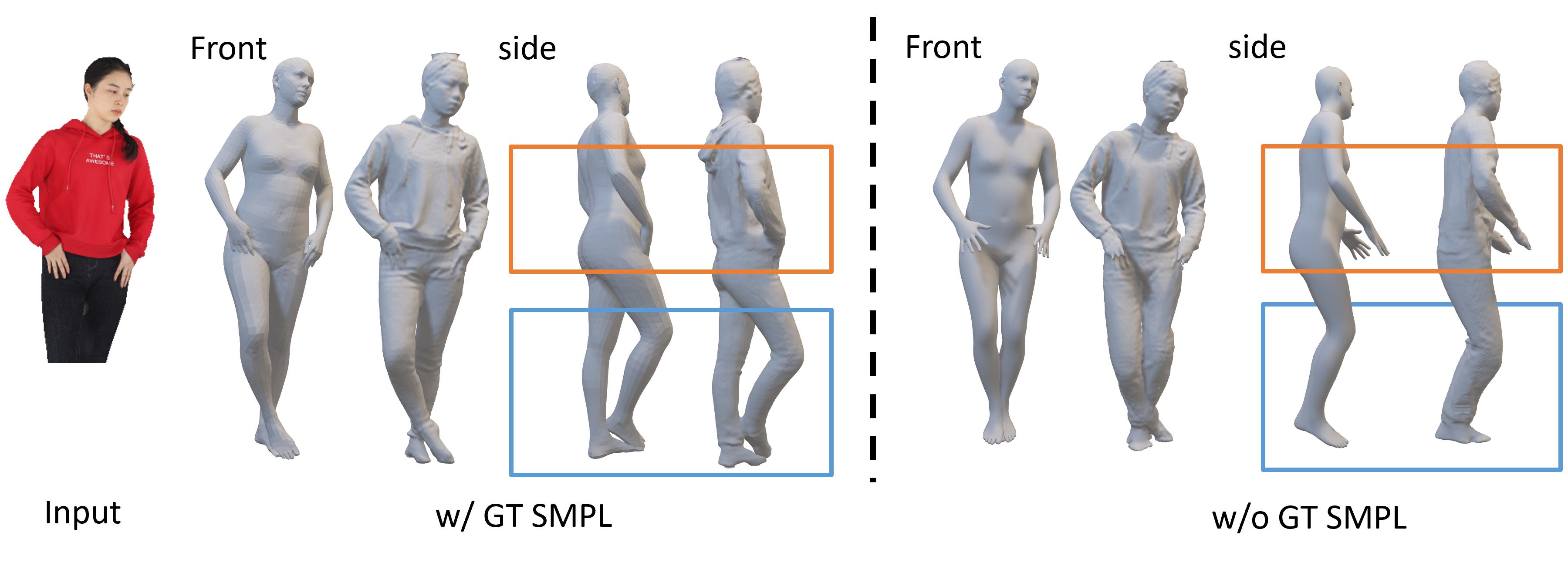}
\caption{Influence of SMPL estimation on reconstruction.}
\end{figure}

\subsubsection{Influence of SMPL Estimation}

Our framework relies on SMPL parameters as pose guidance. 
During training, we employ ground-truth SMPL parameters for stable supervision. 
For fair comparison, the same ground-truth SMPL parameters are also used when evaluating baseline methods that depend on SMPL.

Although SMPL provides useful pose priors, inaccurate SMPL estimation may influence reconstruction quality. 
Figure~\ref{fig:smpl} illustrates examples where SMPL estimation errors lead to imperfect geometry alignment. 
Nevertheless, our method does not require ground-truth SMPL and can operate with SMPL estimates obtained from existing monocular pose estimation methods~\cite{Zhang_2023_ICCV}. 
This demonstrates that our framework remains applicable in practical scenarios where only estimated pose information is available.

\begin{figure*}[ht]
  \centering
   \includegraphics[width=.9\linewidth]{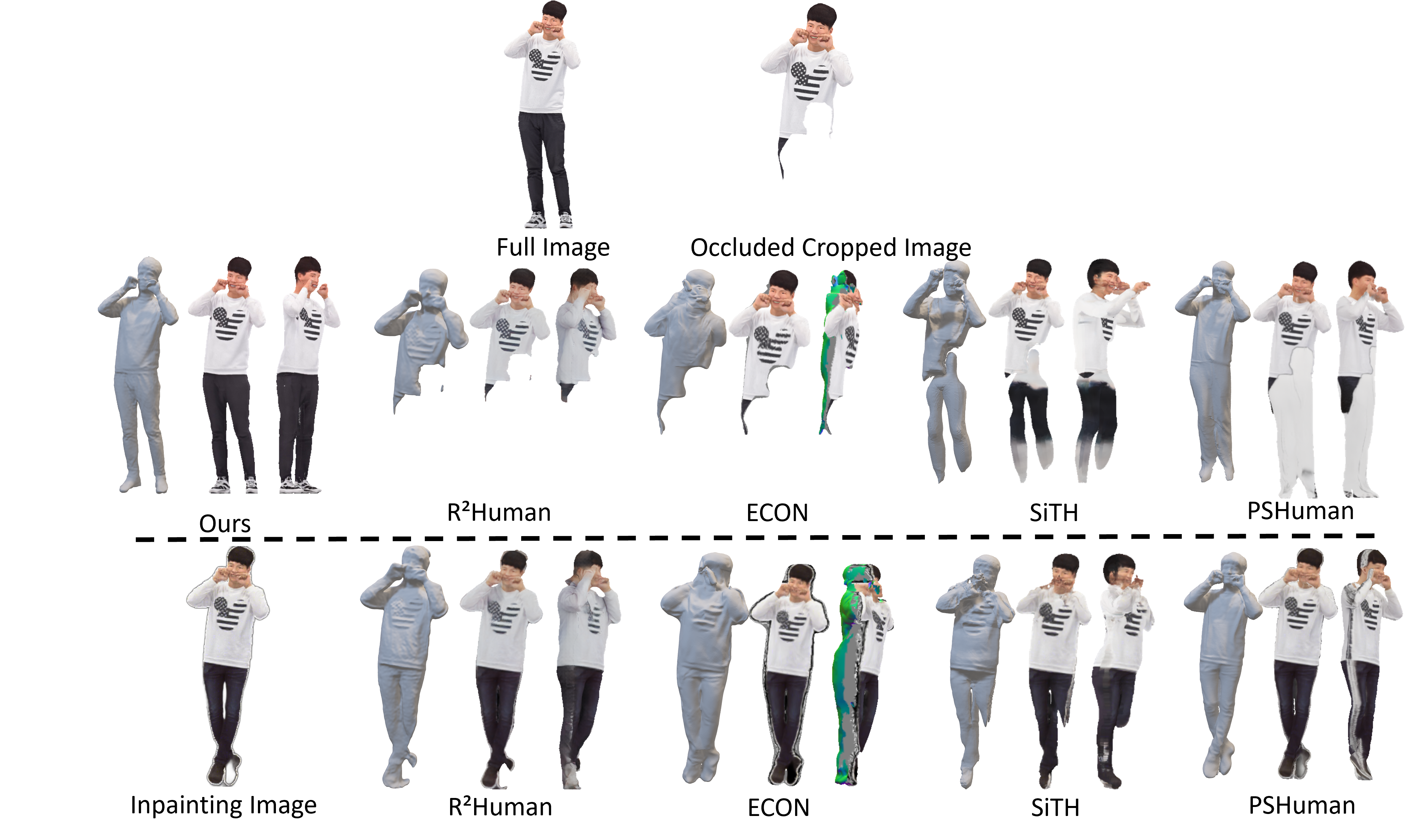}
   \caption{Reconstruction results after 2D inpainting on synthetic data.}
   \label{fig:inp1}
\end{figure*}

\begin{figure*}[ht]
  \centering
   \includegraphics[width=.9\linewidth]{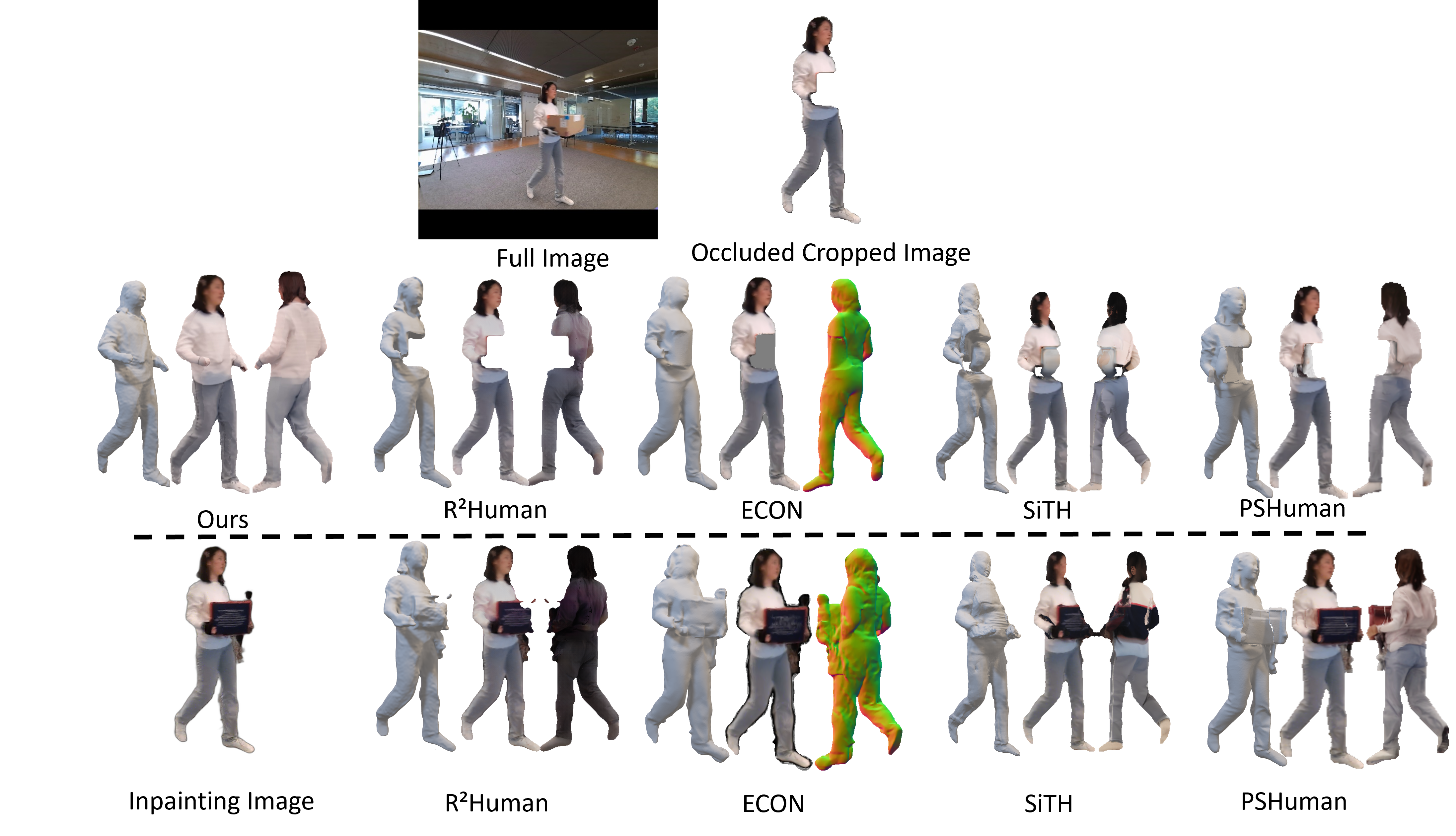}
   \caption{Reconstruction results after 2D inpainting on real-world data.}
   \label{fig:inp2}
\end{figure*}
\begin{table}[h]
    \centering
    \caption{Quantitative comparison on geometry reconstruction and novel view synthesis.}
    \resizebox{.45\textwidth}{!}{%
    \begin{tabular}{c|cccccc}
        \toprule[1.5pt]
        \multirow{2}{*}{Method}
        &\multicolumn{3}{c}{Geometry} &\multicolumn{3}{c}{Novel View Synthesis}\\ 
        \cmidrule(lr){2-4}\cmidrule(lr){5-7}
                    &CD$\downarrow$   &P2S$\downarrow$ &Normal$\downarrow$ &SSIM$\uparrow$   &PSNR$\uparrow$  &LPIPS$\downarrow$\\ 
        \hline
        ECON         &6.2103 &5.7906 &4.5873                 
                     &-      &-      &-      \\
        SiTH         &6.0886 &4.4403 &3.8022                 
                     &0.8830 &18.19  &0.1032 \\
        R${^2}$Human &4.6832 &2.5536 &2.1247                 
                     &0.8762 &17.89  &0.0966\\
        PSHuman      &2.0661 &1.9597 &1.6914
                     &0.9068 &20.79  &0.0645\\
        Ours         
        &\textbf{1.4432} &\textbf{1.3442} &\textbf{1.4613}
        &\textbf{0.9102} &\textbf{22.42 } &\textbf{0.0553}\\
        \bottomrule[1.5pt]
    \end{tabular}%
    }
    \label{tab:4}
\end{table}

\subsection{Comparison with 2D Inpainting + 3D Reconstruction}
An alternative strategy for handling occlusion is to first perform image inpainting and then apply standard monocular 3D reconstruction methods. To investigate this pipeline, we employ the state-of-the-art inpainting model Flux~\cite{blackforestlabs2024flux} to fill occluded regions before reconstruction.

Quantitative results are reported in Table~\ref{tab:4}. Our method directly reconstructs from occluded images, while the baseline pipeline operates on inpainted inputs.

As shown in Figure~\ref{fig:inp1}, inpainting can produce visually plausible results on synthetic data. However, the generated content often introduces geometric inconsistencies that propagate to the reconstruction stage. 

The problem becomes more severe on real-world images (Figure~\ref{fig:inp2}), where occluders frequently interact with the human body. Inpainting often leaves residual artifacts or inaccurate structures, leading to unstable 3D reconstruction. In contrast, our geometry-first pipeline directly infers missing structures in 3D space and produces globally coherent results without relying on perfect 2D completion.

\begin{figure}[t]
  \centering
   \includegraphics[width=1\linewidth]{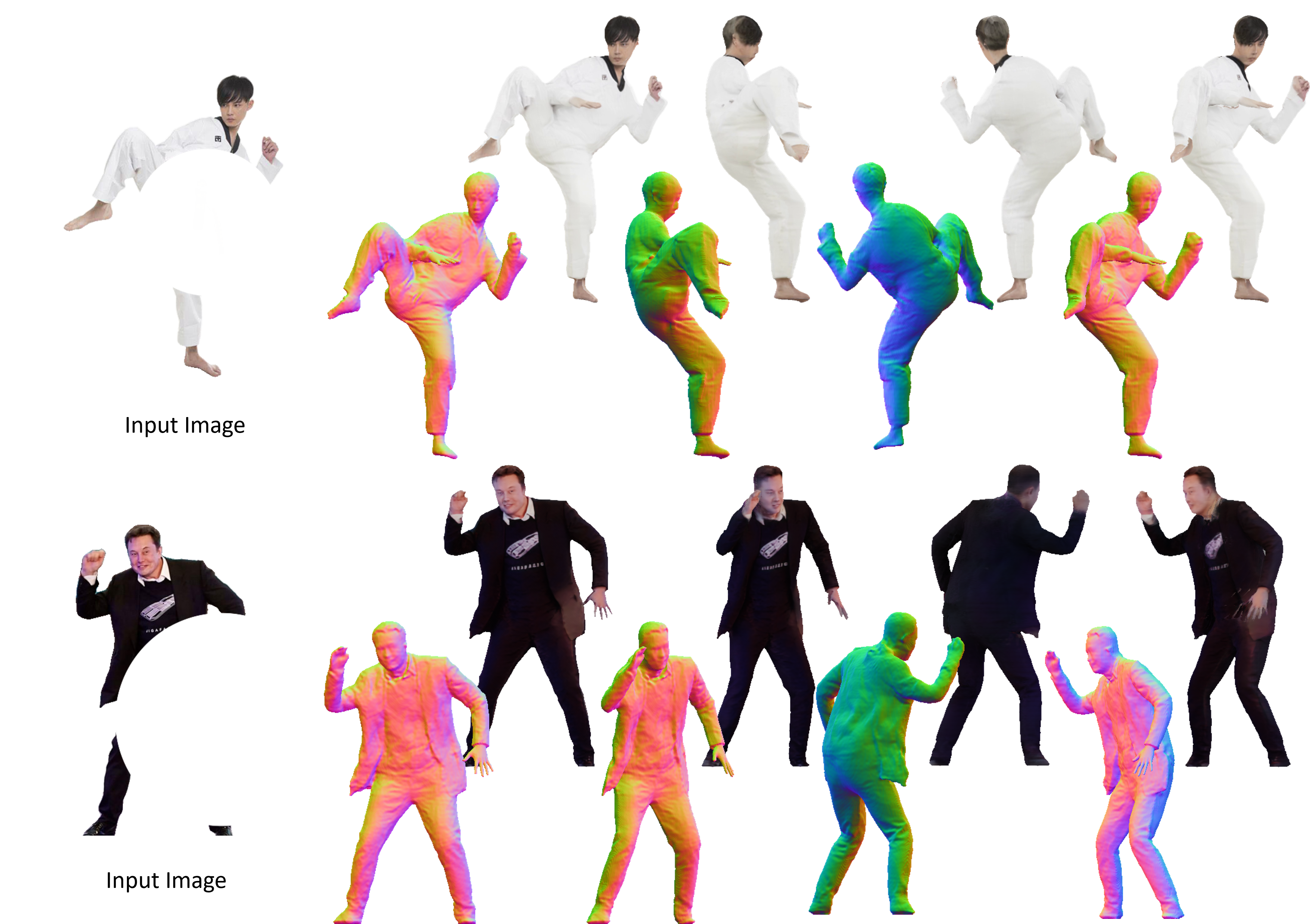}
   \caption{In-the-wild results.}
   \label{fig:wild1}
\end{figure}

\yyw{\subsection{In-the-wild Reconstruction}
We collect images from the Internet for in-the-wild testing. We first use Graphonomy~\cite{gong2019graphonomy} to estimate the human mask in the image, thereby removing the background to obtain the foreground image. We then estimate the SMPL parameters from foreground human images using existing methods~\cite{zhang2021pymaf,Zhang_2023_ICCV}, and use our method to reconstruct a complete and realistic model from a single image. Figure~\ref{fig:wild1} \yl{shows} our in-the-wild reconstruction results, and our method faithfully captures geometry and appearance despite serious occlusion challenges. This demonstrates the robustness and generalization ability of our method, making it suitable for practical deployment in real-world scenarios.}

\section{Conclusion}
We propose a two-stage framework for monocular 3D human reconstruction under occlusion, which explicitly disentangles geometry modeling and texture synthesis. \yyw{This design mitigates the long-standing issue of geometry–texture interference, enabling robust reconstruction even in severely occluded scenarios. 
In the geometry stage, visibility-guided completion with feature-level regularization and dual-view normal refinement enables faithful recovery of human geometry even under severe occlusions. In the texture stage, a geometry-faithful diffusion renderer, conditioned on refined normals and image-adaptive appearance features, synthesizes realistic and view-consistent textures while maintaining strong geometric coherence. }
Extensive experiments demonstrate that our method achieves structurally complete geometry and realistic, view-consistent textures, outperforming state-of-the-art approaches. We believe our framework offers a systematic solution for occlusion-aware human modeling and lays a foundation for applications such as virtual avatars and digital humans.

\yyw{
\textbf{Limitation.} Our method may encounter difficulties when the target person appears extremely small or severely low-resolution in the input image, as limited pixel-level evidence constrains both geometric detail recovery and texture synthesis. Future work will explore improved low-resolution modeling to enhance robustness in such cases.
}

\section{Acknowledgements}
This work was partially supported by the National Key R\&D Program of China (2023YFC3082100), the National Natural Science Foundation of China (62371013) and the Science Fund for Distinguished Young Scholars of Tianjin under Grant (22JCJQJC00040).

\bibliographystyle{IEEEtran}
\bibliography{IEEEabrv, citations}

\vfill


\end{document}